\documentclass{article} 
\usepackage[preprint]{colm2026_conference}

\usepackage{microtype}
\usepackage{hyperref}
\usepackage{url}
\usepackage{booktabs}

\usepackage{amsmath,amssymb}
\usepackage[most]{tcolorbox}
\usepackage{lmodern}
\usepackage{graphicx}
\usepackage{booktabs}
\usepackage{multirow}
\usepackage{array}
\usepackage[table]{xcolor}
\usepackage{algorithm}
\usepackage{algpseudocode}
\usepackage{hyperref}
\usepackage{enumitem}
\usepackage{comment}
\usepackage{subcaption}
\usepackage{wrapfig}

\usepackage{lineno}
\newcommand{\ours}{AriadneMem}

\definecolor{darkblue}{rgb}{0, 0, 0.5}
\hypersetup{colorlinks=true, citecolor=darkblue, linkcolor=darkblue, urlcolor=darkblue}

\title{AriadneMem: Threading the Maze of Lifelong Memory for LLM Agents}

\author{\textbf{Wenhui Zhu}$^{1}$\thanks{Equal contribution} ,
  \textbf{Xiwen Chen}$^{2}$\footnotemark[1] ,
  \textbf{Zhipeng Wang}$^{3}$\footnotemark[1] , 
  \textbf{Jingjing Wang}$^{4}$, 
  \textbf{Xuanzhao Dong}$^{1}$, \\
  \textbf{Minzhou Huang}$^{5}$, 
   \textbf{Rui Cai}$^{6}$, 
  \textbf{Hejian Sang}$^{7}$,
  \textbf{Hao Wang}$^{4}$,
  \textbf{Peijie Qiu}$^{8}$, \\
  \textbf{Yueyue Deng}$^{9}$, 
  \textbf{Prayag Tiwari}$^{10}$, 
  \textbf{Brendan Hogan Rappazzo}$^{2}$, 
   \textbf{Yalin Wang}$^{1}$
  \\
  \\
  $^1$Arizona State University, $^2$Morgan Stanley, $^3$Rice University, 
   $^4$Clemson University, \\ $^5$Northwestern University, $^6$UC Davis, $^7$Iowa State University, \\
  $^8$Washington University in St. Louis, 
  $^9$Columbia University, $^{10}$Halmstad University \\
  \texttt{wzhu59@asu.edu, xiwen.chen@morganstanley.com} \\
}

\begin{document}

\ifcolmsubmission
\linenumbers
\fi

\maketitle



\begin{abstract}
Long-horizon LLM agents require memory systems that remain accurate under fixed context budgets.
However, existing systems struggle with two persistent challenges in long-term dialogue: (i) \textbf{disconnected evidence}, where multi-hop answers require linking facts distributed across time, and (ii) \textbf{state updates}, where evolving information (e.g., schedule changes) creates conflicts with older static logs.
We propose \textbf{\ours{}}, a structured memory system that addresses these failure modes via a decoupled two-phase pipeline.
In the \textbf{offline construction phase}, \ours{} employs \emph{entropy-aware gating} to filter noise and low-information message before LLM extraction and applies \emph{conflict-aware coarsening} to merge static duplicates while preserving state transitions as temporal edges.
In the \textbf{online reasoning phase}, rather than relying on expensive iterative planning, \ours{} executes \emph{algorithmic bridge discovery} to reconstruct missing logical paths between retrieved facts, followed by \emph{single-call topology-aware synthesis}.
On LoCoMo experiments with GPT-4o, \ours{} improves \textbf{Multi-Hop F1 by 15.2\%} and \textbf{Average F1 by 9.0\%} over strong baselines. Crucially, by offloading reasoning to the graph layer, \ours{} reduces \textbf{total runtime by 77.8\%} using only \textbf{497} context tokens. The code is available at \url{https://github.com/LLM-VLM-GSL/AriadneMem}.
\end{abstract}

\section{Introduction}

\begin{wrapfigure}{r}{0.37\linewidth}
  \vspace{-6pt}
  \centering
  \includegraphics[width=\linewidth]{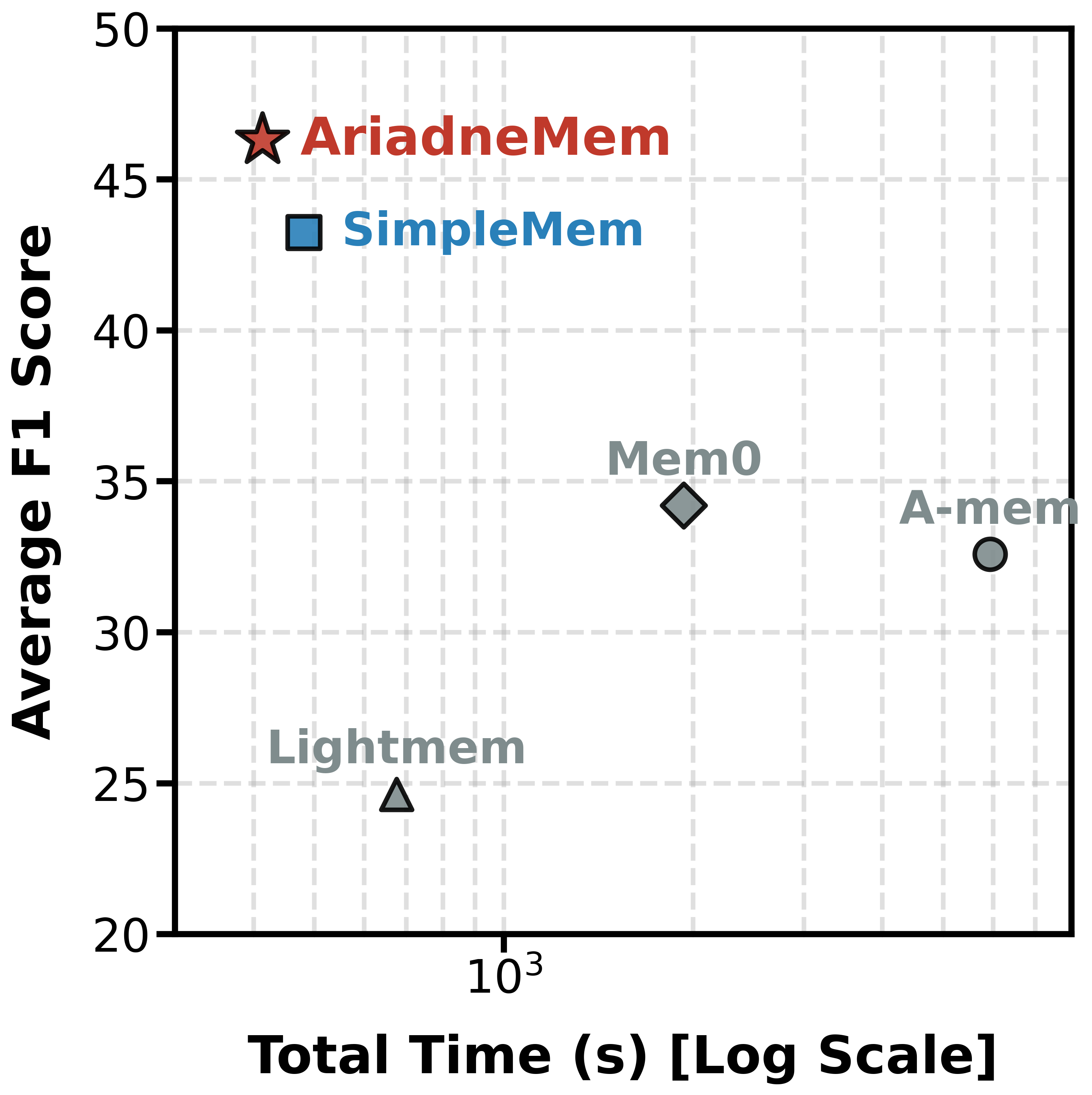}
  \vspace{-0.2in}
  \caption{Efficiency-Performance Trade-off on LoCoMo benchmark.}
  \vspace{-0.4in}
  \label{fig:trade-off}
\end{wrapfigure}
Large Language Model (LLM) agents operating in persistent, open-ended environments require robust long-term memory to maintain state consistency and perform multi-step reasoning over time. While large context windows have expanded, they remain insufficient for lifelong interaction and suffer from "lost-in-the-middle" phenomena~\cite{liu2023lost}. 
Consequently, external memory systems have become a standard component of agent architecture, typically following a Retrieve-Augmented Generation (RAG) paradigm~\cite{rag1}. However, within this paradigm, existing approaches face a fundamental trade-off between \textit{structure} (how efficiently information is stored) and \textit{connectivity} (how effectively scattered facts are linked).


Prior works navigate this trade-off through diverging strategies. Some systems rely on (i) raw-log retrieval~\cite{rag1}, which preserves context but lacks structural connectivity. To bridge gaps in retrieval, others employ (ii) iterative reasoning~\cite{du2025memr}, "simulating" connectivity by generating multiple queries and performing repeated search loops. While effective, this iterative process incurs significant latency and token costs~\cite{Packer2023MemGPTTL}. Representing the current state-of-the-art in storage efficiency, SimpleMem~\cite{liu2026simplemem} shifts the paradigm by compressing dialogues into context-independent atomic entries. These entries function as self-contained semantic units where pronominal ambiguity is resolved and relative time is grounded to absolute timestamps (e.g., converting "He'll meet Bob tomorrow" to "Alice meets Bob on 2025-11-16"). This maximizes information density and reduces ambiguity.

Crucially, however, SimpleMem occupies a precarious middle ground.
While it significantly improves on the information density of Strategy (i), it inadvertently retains its topological flatness, failing to resolve the challenge of \textit{disconnected evidence}.
Because it treats memory as a set of isolated atoms without intrinsic links, SimpleMem is forced to fall back on the expensive planning loops of Strategy (ii) to perform multi-hop reasoning (e.g., $A \rightarrow B \rightarrow C$), invoking the LLM to deduce intermediate "bridge" nodes ($B$) that lack direct semantic overlap.

Consequently, a distinct paradox emerges: While SimpleMem is highly efficient in terms of storage tokens, it suffers from significant interaction latency for complex queries.
The reliance on runtime LLM inference to \textit{bridge} disjoint facts effectively negates its speed advantage during multi-hop reasoning, introducing inference overhead precisely when the agent needs to think deepest.

Compounding this latency is a fundamental fragility regarding \textit{state updates}:
Inheriting the limitations of Strategy (i), the system lacks explicit temporal structure, struggling to distinguish between redundant repetition and evolving information (e.g., ``meeting at 2pm'' $\rightarrow$ ``changed to 3pm'').

\begin{figure}
    \centering
    \includegraphics[width=0.9\linewidth]{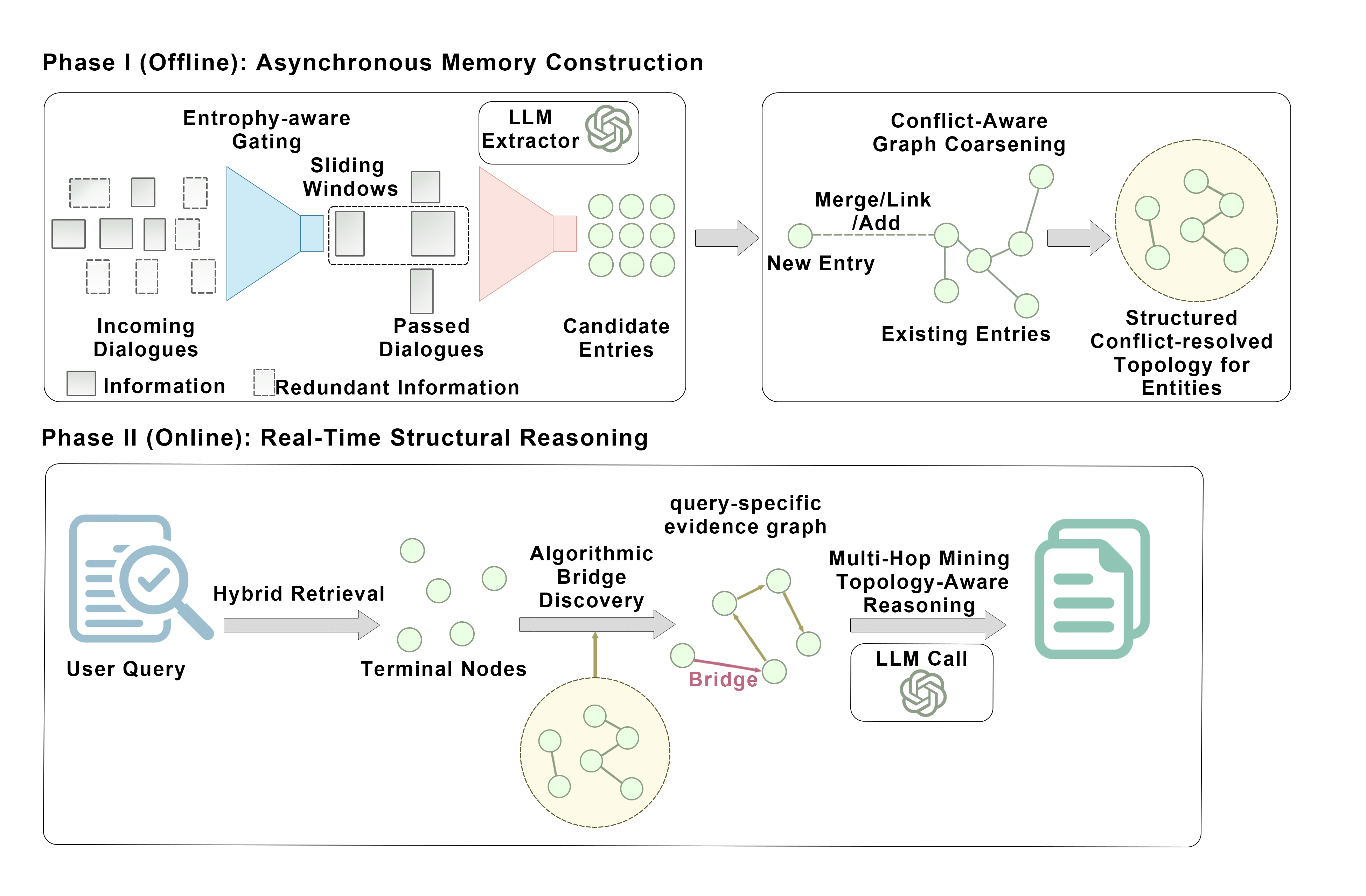}
    \caption{\textbf{Overview of the AriadneMem architecture.} The pipeline is decoupled into two phases: (I) Offline Memory Construction, which maintains an evolutionary graph via entropy-aware gating and conflict-aware coarsening to resolve state updates; and (II) Online Structural Reasoning, which connects disjoint evidence through algorithmic bridge discovery and performs topology-aware synthesis.}
    \label{fig:overview}
\end{figure}

In this work, we argue that the limitation lies in the \textit{representation} itself. Relying on the LLM to implicitly reconstruct logical chains from flat fragments is both computationally expensive and prone to error. The key observation is that complex questions are rarely answered by isolated facts, but by connected \emph{chains} of memories. To simultaneously eliminate retrieval-time latency and ground multi-hop inference in explicit topology, long-term memory should not be a flat set, but an evolutionary graph that directly encodes state transitions and entity relationships. We introduce \textbf{\ours{}}\footnote{Named after the Greek mythological figure Ariadne, who gave Theseus a thread to navigate the Labyrinth. Analogously, our system provides a structural \textit{thread} (reasoning path) for agents to navigate the complex maze of lifelong memory.} (overview of the framework is presented in Fig.~\ref{fig:overview}), a framework that transforms memory retrieval from a probabilistic guessing game into a deterministic structural traversal.

Specifically, \ours{} employs entropy-aware graph coarsening to manage memory density, merging semantically redundant nodes while strictly preserving state updates. Crucially, leveraging this graph structure enables us to execute multi-hop planning via Approximate Steiner Tree retrieval. Instead of a flat list, this process constructs a compact \emph{query-specific evidence graph}, automatically discovering ``bridge nodes'' to connect scattered evidence. Within this subgraph, we mine explicit multi-hop paths (via DFS) which are serialized to guide the generator. This structured context enables \ours{} to perform a single LLM call for final answer synthesis, thereby circumventing the expensive iterative planning loops required by prior methods.

Our contributions are summarized as follows:
\begin{itemize}[leftmargin=1.5em]
    \item \textbf{From Iterative Planning to Structural Traversal:} We identify a fundamental efficiency bottleneck in existing agents: the reliance on expensive LLM reasoning to bridge disjoint memories. We propose to shift this burden to a graph-native layer, utilizing \textit{algorithmic bridge discovery} and \textit{bounded-depth path mining} to deterministically reconstruct evidence chains. This transition reduces interaction latency by \textbf{77.8\%} while elevating multi-hop reasoning accuracy.

    \item \textbf{Entropy-Aware Evolutionary Memory:} Unlike static vector stores that suffer from information redundancy or catastrophic forgetting of state updates, \ours{} maintains an evolving memory graph. By introducing \textit{conflict-aware coarsening}, we merge redundant semantics while explicitly encoding state transitions as temporal links, ensuring the agent maintains a consistent "world model" of the dialogue history.

    \item \textbf{Topology-Aware Contextualization:} We present a novel serialization paradigm that injects the structural properties of the retrieved subgraph directly into the LLM. By providing \textit{path-oriented grounding} rather than a flat list of fragments, \ours{} effectively mitigates the "lost-in-the-middle" phenomenon and ensures high-fidelity answer synthesis within a compact context budget (avg. \textbf{497} tokens).

    \item \textbf{Superior Performance:} Extensive experiments on the LoCoMo benchmark demonstrate that \ours{} achieves a significant leap in both reasoning quality and operational efficiency. We report a \textbf{15.2\% improvement in Multi-Hop F1} and a \textbf{9.0\% gain in Average F1} over the current SOTA, establishing \ours{} as a highly practical framework for lifelong LLM agents. See Fig.~\ref{fig:trade-off} for performance-efficiency trade-off across different models.
\end{itemize}

\section{Method}
\label{sec:method}

\subsection{Problem Setup \& Pipeline Overview}
\label{sec:overview}
We consider a stream of dialogues $D=\{d_t\}_{t=1}^T$ where each item is $d_t=\langle s_t,x_t,t_t\rangle$ (speaker, text, optional timestamp).
The system stores \emph{atomic entries} $\mathcal{M}=\{m_k\}$, each with a lossless restatement $S_k$, keyword set $\mathcal{K}_k$, metadata record $\mathcal{R}_k$ (persons/entities/location/time)~\cite{liu2026simplemem}, and dense embedding vector $\mathbf{v}_k\in\mathbb{R}^d$.
Entries are indexed in a multi-view store: a semantic (dense) index for similarity search and a lexical (sparse) index for keyword matching.
Given a user query $q$, the system retrieves a subgraph $G_q \subset \mathcal{M}$ to synthesize the final answer $a$:
\begin{equation}
    a = \mathrm{LLM}\big(q, \mathrm{Serialize}(G_q)\big).
\end{equation}

\ours{} implements a pipeline decoupled into two asynchronous phases, aligned with Figure~\ref{fig:overview}:
\begin{itemize}[leftmargin=1.2em]
    \item \textbf{Phase I (Offline): Asynchronous Memory Construction.} 
    To process the continuous stream efficiently, incoming dialogues are subject to \textit{entropy-aware gating} to filter low-information inputs before passing through \textit{atomic extraction} to produce structured entries. The resulting facts are then \textit{coarsened} into an evolutionary graph, where the system merges redundant duplicates while preserving state updates (e.g., changing schedules) as explicit temporal links.

    \item \textbf{Phase II (Online): Real-Time Structural Reasoning.} 
    Upon receiving a query, we first check \textit{fast paths} (cache and regex lookup) for immediate answers. If unresolved, we perform \textit{hybrid retrieval} to find entry points, followed by \textit{algorithmic bridge discovery} to connect disjoint evidence chains into a \textbf{query-specific evidence graph}. Finally, we perform \textit{single-call topology-aware synthesis}, serializing this graph to guide the LLM in generating a multi-hop answer in one pass.
\end{itemize}

\subsection{Phase I: Asynchronous Memory Construction}
\label{sec:phase1}
This phase transforms the raw stream $D$ into a sparse, conflict-resolved evolutionary graph.
We formally define an atomic entry $m \in \mathcal{V}$ as a tuple $m = \langle \mathbf{v}, \mathcal{K}, \mathrm{Ent}, t \rangle$, containing a dense embedding $\mathbf{v} \in \mathbb{R}^d$, a keyword set $\mathcal{K}$, extracted entities $\mathrm{Ent}$, and timestamp $t$.

\paragraph{Entropy-Aware Gating.}
To prevent the memory store from being flooded with trivial chit-chat, we employ a pre-extraction gating mechanism.
Let $E(\cdot)$ be the embedding function. For an incoming dialogue $d_t$, we retrieve its nearest neighbor $m^*(d_t)$ in the existing memory and compute a redundancy score $r_t$:
\begin{equation}
    m^*(d_t) = \operatorname*{arg\,max}_{m \in \mathcal{V}} \cos\big(E(x_t), \mathbf{v}_m\big), \quad r_t = \cos\big(E(x_t), \mathbf{v}_{m^*}\big).
\end{equation}
We apply a non-linear gating decision $\Phi_{\text{gate}}(d_t)$ that blocks short-term repetition but allows long-term recurrence:
\begin{equation}
    \Phi_{\text{gate}}(d_t)=
    \begin{cases}
        0, & r_t > \lambda_{\text{red}} \ \wedge\ \Delta(t_t, t_{m^*}) < \delta_{\text{short}}, \\
        1, & \text{otherwise},
    \end{cases}
\end{equation}
where $\lambda_{\text{red}}$ is the redundancy threshold and $\delta_{\text{short}}$ is a short-horizon window (e.g., 1 hour) to filter near-immediate repetitions. If $\Phi_{\text{gate}}(d_t)=0$, we drop the input immediately.

\paragraph{Atomic Entry Extraction.}
For dialogues that pass gating ($\Phi=1$), we invoke the LLM extractor $\mathcal{F}_{\theta}$ over a sliding window $W_t$.
\begin{equation}
    \{m_k\} = \mathcal{F}_{\theta}(W_t).
\end{equation}
This step produces a set of candidate atomic entries. By placing the gating mechanism \emph{before} extraction, \ours{} reduces redundant extraction calls compared to pipelines that extract from every dialogue turn.

\paragraph{Conflict-Aware Graph Coarsening.}
Simply appending new entries leads to linear growth in storage. We coarsen the candidate entries to maintain a sparse topology by distinguishing \emph{redundancy} from \emph{state updates}.
For a new entry $m$ and an existing entry $\tilde{m}$, we compute semantic similarity ($\mathrm{sim}$) and keyword overlap ($\mathrm{ovlp}$):
\begin{equation}
    \mathrm{sim}(m,\tilde{m})=\cos(\mathbf{v}_m,\mathbf{v}_{\tilde{m}}), \quad
    \mathrm{ovlp}(m,\tilde{m})=\frac{|\mathcal{K}_m \cap \mathcal{K}_{\tilde{m}}|}{\max(1,|\mathcal{K}_m|)}.
\end{equation}
We classify the relationship into three distinct actions using thresholds $\lambda_{\text{coal}}$ and $\lambda_{\text{ovlp}}$:
\begin{equation}
    \text{Action} = 
    \begin{cases} 
        \textsc{Merge}, & \text{if } \mathrm{sim} > \lambda_{\text{coal}} \land \mathrm{ovlp} > \lambda_{\text{ovlp}} \quad (\text{Duplicate}) \\
        \textsc{Link}, & \text{if } \mathrm{sim} > \lambda_{\text{coal}} \land \mathrm{ovlp} \le \lambda_{\text{ovlp}} \quad (\text{State Update}) \\
        \textsc{Add}, & \text{otherwise}.
    \end{cases}
\end{equation}
Specifically, \textbf{\textsc{Merge}} discards true duplicates and updates the timestamp of $\tilde{m}$.
Crucially, \textbf{\textsc{Link}} handles cases where semantics align but details differ (e.g., "meeting at 2pm" vs "meeting at 3pm"); here, we retain $m$ and create a directed temporal edge $\tilde{m} \to m$, explicitly preserving the state transition fidelity.
Finally, \textbf{\textsc{Add}} inserts distinct inputs as new isolated nodes.

\paragraph{Outcome: Graph as a Navigational Prior.}
By the end of Phase I, the raw stream has been transformed into a structured, conflict-resolved topology. This offline construction serves a dual purpose for the online phase:
(1) \textbf{Bridge Enablement:} The established temporal and entity links provide the necessary "roads" for the Steiner Tree algorithm (Phase II) to discover hidden connections between disjoint facts; and
(2) \textbf{State Resolution:} By explicitly modeling updates as directed edges (e.g., $2\text{pm} \to 3\text{pm}$), the graph resolves potential paradoxes before retrieval, ensuring the agent acts on the latest state without needing to reason over conflicting raw logs.

\subsection{Phase II: Real-Time Structural Reasoning}
\label{sec:phase2}
Upon receiving a query $q$, \ours{} executes a synchronous pipeline to construct a query-specific evidence graph $G_q$ and synthesize an answer $a$. We reformulate this process as an algorithmic search problem rather than a generative planning task.

\paragraph{Fast Paths (Heuristic Short-Circuiting).}
Before launching general-purpose retrieval, we check two lightweight shortcuts that do not require additional LLM calls:
(i) an \emph{enhanced cache} for common query patterns built during ingestion (e.g., maintaining a running counter for ``How many emails from Alice?''), and
(ii) a regex-based \emph{attribute lookup} for simple ``X's attribute'' questions (e.g., directly mapping ``What is Bob's phone number?'' to metadata fields).
If either path returns sufficient evidence, the system directly constructs a minimal graph containing the lookup facts, bypassing the heavier retrieval steps.

\paragraph{Hybrid Retrieval.}
For complex queries, we first identify a set of \textit{terminal nodes} $V_{\text{term}}$ via hybrid retrieval (dense + lexical):
\begin{equation}
    V_{\text{term}} = \mathrm{Top}\text{-}k_{\text{sem}}(q) \ \cup\ \mathrm{Top}\text{-}k_{\text{lex}}(q).
\end{equation}
Consistent with multi-view retrieval principles, we use $E(q)$ for dense similarity and keyword matching for lexical retrieval. We treat $k_{\text{sem}}$ and $k_{\text{lex}}$ as hyperparameters. Additionally, we extract target entities from $q$ using simple heuristics to downweight candidates that lack entity alignment.

\paragraph{Base Graph Construction.}
We construct a base graph $G_0=(V_{\text{term}}, E_0)$ to establish initial connectivity for the query-specific evidence graph. Let $\mathrm{Ent}(m)$ denote the set of entities extracted for node $m$. We define a directed edge $m_i \to m_j$ if nodes share entities or are temporally close:
\begin{equation}
    \mathrm{Ent}(m_i) \cap \mathrm{Ent}(m_j) \neq \varnothing \quad \text{or} \quad |t_i - t_j| < \delta_{\text{time}}.
\end{equation}
In our reference implementation, we set $\delta_{\text{time}} = 6 \text{ hours}$. This strict window encourages a "narrative backbone," linking events that likely belong to the same immediate context while keeping the graph sparse.

\paragraph{Algorithmic Bridge Discovery (Steiner Tree Approximation).}
Terminal nodes are often topologically disconnected. To recover these links without retrieval-time planning loops, we approximate a Steiner Tree by searching for \textit{bridge nodes} $b^*$.
For disconnected pairs $(m_i, m_j) \in V_{\text{term}}$, we query the memory for a node that maximizes semantic connectivity within the valid time interval:
\begin{equation}
    b^* = \operatorname*{arg\,max}_{m \in \mathcal{V} \setminus V_{\text{term}}} \cos\big(E(\mathbf{q}_{ij}), \mathbf{v}_m\big)
    \ \ \text{s.t.}\ \ t_m \in [\min(t_i,t_j), \max(t_i,t_j)].
\end{equation}
Here, the bridge query $\mathbf{q}_{ij}$ is constructed by concatenating the entities and keywords of the endpoints: $\mathrm{Ent}(m_i) \cup \mathrm{Ent}(m_j) \cup \mathcal{K}_i \cup \mathcal{K}_j$.
To maintain precision, we attempt bridge search only when the time gap is moderate (1--168 hours) and strictly consider only the top-5 candidates from semantic search. If a valid bridge is found, it is added to the graph ($m_i \to b^* \to m_j$).

\paragraph{Multi-Hop Path Mining \& Node Budget.}
To provide structured guidance, we explicitly mine reasoning paths via Depth-First Search (DFS) on the augmented graph $G_q$.
Let $\mathcal{P}_q$ be the set of directed paths up to length $L$ (set to $L=3$ hops in our implementation):
\begin{equation}
    \mathcal{P}_q = \{p \in G_q \mid 2 \le |p| \le L, \text{temporally consistent}\}.
\end{equation}
To control the context length for generation, we enforce a \textbf{node budget} (e.g., keeping between 8 and 25 nodes). We prioritize paths based on length and temporal coherence, pruning the graph if the budget is exceeded.

\paragraph{Topology-Aware Reasoning.}
Finally, we generate the answer by conditioning a single LLM call on the serialized topology.
We serialize $G_q$ and $\mathcal{P}_q$ into a textual format $\mathcal{C}_{\text{graph}}$, listing timestamped facts and explicit path indicators.
Crucially, to ensure robustness on benchmarks like LoCoMo, we attach a set of \textbf{explicit answer rules}:
(i) \textit{Length Constraints} (concise vs. list based on query type);
(ii) \textit{Temporal Fidelity} (strict copying of timestamps, avoiding relative normalization);
(iii) \textit{Aggregation Logic} (specifying counting formats); and
(iv) \textit{Formatting} (JSON requirements as output).
The final answer is synthesized via:
\begin{equation}
    a = \mathrm{LLM}(q, \mathcal{C}_{\text{graph}}).
\end{equation}
This structured context enables \ours{} to perform complex multi-hop reasoning in a single inference step, circumventing the latency of iterative generation. A qualitative example is presented in Fig.~\ref{fig:example}.

\begin{algorithm}[h]
\caption{\ours{} Pipeline: Asynchronous Memory \& Reasoning}
\label{alg:method_pipeline}
\begin{algorithmic}[1]
\Require Stream $D$, Memory $\mathcal{V}$, Index $\mathbb{I}$, Thresholds $\lambda_{\{\text{red,coal,ovlp}\}}$, Windows $\delta_{\{\text{short,time}\}}$
\Ensure Answer $a$ for query $q$

\Statex \textbf{\textsc{Phase I: Asynchronous Memory (Offline)}}
\For{$d_t \in D$}
    \State $m^* \leftarrow \operatorname{NN}(x_t, \mathcal{V})$; \quad $r_t \leftarrow \cos(E(x_t), \mathbf{v}_{m^*})$
    \State \textbf{if} $r_t > \lambda_{\text{red}} \land \Delta t < \delta_{\text{short}}$ \textbf{then} \textbf{continue} \Comment{Gating $\Phi_{\text{gate}}$}
    \State Buffer $d_t$ into $W_t$. \textbf{if} full:
        \State \quad $\{m\} \leftarrow \mathcal{F}_{\theta}(W_t)$; \textbf{for} each $m$ vs $\tilde{m} \in \mathcal{V}$:
        \State \quad \quad $\text{Op} \leftarrow (\text{sim} > \lambda_{\text{coal}}) ?\ (\text{ovlp} > \lambda_{\text{ovlp}} ? \textsc{Merge} : \textsc{Link}) : \textsc{Add}$
        \State \quad \quad Execute Op (Update $\mathcal{V}, \mathbb{I}$; Link $\tilde{m}\to m$ if Link); Reset $W_t$
\EndFor

\Statex \textbf{\textsc{Phase II: Structural Reasoning (Online)}}
\State $V_{\text{term}} \leftarrow \mathrm{Top}\text{-}k_{\text{sem}}(q, \mathbb{I}) \cup \mathrm{Top}\text{-}k_{\text{lex}}(q, \mathbb{I})$
\State $G_q \leftarrow (V_{\text{term}}, E)$ where $E = \{(u,v) \mid \mathrm{Ent}_u \cap \mathrm{Ent}_v \neq \varnothing \lor \Delta t < \delta_{\text{time}}\}$
\For{disconnected $(m_i, m_j) \in G_q$}
    \State $\mathbf{q}_{ij} \leftarrow \mathrm{Ent}_i \cup \mathrm{Ent}_j \cup \mathcal{K}_i \cup \mathcal{K}_j$
    \State $b^* \leftarrow \operatorname*{arg\,max}_{m \in \mathcal{V} \setminus V_{\text{term}}} \cos(E(\mathbf{q}_{ij}), \mathbf{v}_m)$ s.t. $t_m \in [t_i, t_j]$
    \State \textbf{if} $b^*$ found \textbf{then} $G_q \leftarrow G_q \cup \{m_i \to b^* \to m_j\}$ \Comment{Steiner Approx.}
\EndFor
\State $\mathcal{C}_{\text{graph}} \leftarrow \mathrm{Serialize}(\mathrm{DFS}(G_q, L))$ s.t. Node Budget
\State \Return $\mathrm{LLM}(q, \mathcal{C}_{\text{graph}})$
\end{algorithmic}
\end{algorithm}

\begin{figure*}[!t]
\begin{tcolorbox}[colback=blue!5!white, colframe=blue!8!black, title=Question: What musical artists/bands has Melanie seen? \\ \textcolor{yellow}{\textbf{[Ground Truth: Summer Sounds, Matt Patterson]}}, fonttitle=\bfseries]

\textbf{Facts:}

\textbf{...}

\textbf{...}









\textbf{[F11]} 2023-07-20: Melanie and her family went to the beach recently and the kids had a blast.

\textbf{[F12] }2023-08-13: Melanie celebrated her daughter's birthday with a concert by Matt Patterson.

\textbf{[F13]} 2023-08-24: Melanie spent the day with her family volunteering at a homeless shelter yesterday.

\textbf{[F14]} 2023-08-27: Melanie took her kids to a park yesterday.

\textbf{[F15] }2023-08-28: Melanie mentioned she is a fan of both classical music like Bach and Mozart, as well as modern music like Ed Sheeran's 'Perfect'

\textbf{[F16]} 2023-08-28: Melanie stated that she plays the clarinet and started when she was young.

\textbf{[F17]} 2023-08-28: Melanie mentioned that the band 'Summer Sounds' played an awesome pop song that got everyone dancing and singing.

\textbf{...}

\textbf{...}
\\

\textbf{Reasoning Path:}
\textbf{...}

\textbf{...}

\textbf{F3   }  → \textbf{F18} → \textbf{F19}

\textbf{F18} → \textbf{F19} → \textbf{F20}

\textbf{F15} → \textbf{F16} → \textbf{F17}
  
\textbf{...}

  
  
  
  


\textbf{Final Answer:}
\[
\textbf{Matt Patterson, Summer Sounds}
\]
\end{tcolorbox}
\caption{\textbf{Qualitative Example of Structural Reasoning.} A sample output showing how AriadneMem retrieves and serializes a coherent, timestamped evidence chain to answer a multi-hop question.}
\label{fig:example}
\end{figure*}

\section{Related Work}
\label{sec:related}
\textbf{Memory Systems for LLM Agents.}
Recent approaches manage memory through virtual context or structured representations.
Virtual context methods such as MemGPT~\cite{Packer2023MemGPTTL} provide paging and controller-style memory, but often store raw logs, inducing redundancy and increasing processing cost.
Structured memory systems such as Mem0~\cite{mem0,Chhikara2025Mem0BP}, A-Mem~\cite{Xu2025AMEMAM}, and LightMem~\cite{fang2025lightmem} improve coherence and retrieval but can still preserve referential and temporal ambiguities if the stored text remains minimally processed.
SimpleMem~\cite{liu2026simplemem} addresses this by semantic structured compression into context-independent atomic entries, plus query-aware retrieval to improve token efficiency.
Compared to structured-memory baselines such as SimpleMem~\cite{liu2026simplemem}, \ours{} focuses on constructing a connected evidence subgraph rather than returning an unstructured top-$k$ list.

\textbf{Context Management and Retrieval Efficiency.}
Beyond storage, efficient access to historical information is a core challenge.
Retrieval-augmented generation (RAG)~\cite{rag1} decouples memory from inference, but flat top-$k$ retrieval can miss intermediate evidence required by multi-hop questions.
Graph-based RAG variants (e.g., GraphRAG~\cite{graphrag}) build structured summaries for query-focused retrieval, mainly targeting static knowledge corpora.
In long-term episodic memory, evidence must additionally preserve temporal flow and state updates.
\ours{} addresses these challenges by (i) conflict-aware coarsening for updates and (ii) approximate Steiner completion with bridge-node discovery to construct connected evidence subgraphs.

\textbf{Compression and Token Efficiency.}
Prompt compression methods (e.g., LLMLingua~\cite{pan2024llmlingua}) reduce token usage but may lose task-critical details when applied post hoc.
SimpleMem~\cite{liu2026simplemem} emphasizes write-time semantic lossless compression, ensuring each memory entry is self-contained.
\ours{} focuses on retrieval-time structure: rather than expanding to large contexts or running iterative planning loops, we retrieve a compact connected subgraph and expose explicit reasoning paths to the generator.

\section{Experiments}
\label{sec:experiments}
We evaluate \ours{} on long-term conversational memory benchmarks such as LoCoMo~\cite{locomo} to answer three questions. First, does \ours{} improve long-horizon QA accuracy compared to prior agent memory systems, with emphasis on MultiHop and Temporal subsets. Second, does it improve efficiency when we measure both retrieved context length and runtime, including retrieval time and end-to-end time. Third, which components account for the gains, as measured by ablation and retrieval-depth sensitivity.
Following the reporting style of SimpleMem~\cite{liu2026simplemem}, we report per-subset F1 and BLEU together with Token Cost in Table~\ref{tab:high_cap_results}, runtime breakdown in Table~\ref{tab:efficiency}, and ablations and sensitivity in Tables~\ref{tab:ablation_by_type} and~\ref{tab:combined_k_sensitivity}.

\subsection{Implementation Details}
\label{sec:impl_details}
We follow the LoCoMo evaluation protocol and report results on MultiHop, Temporal, OpenDomain, and SingleHop subsets. Within each block of Table~\ref{tab:high_cap_results}, all methods are evaluated under the same underlying LLM backbone. We report F1 and BLEU for answer quality. We also report token cost, defined as the number of tokens in the retrieved memory context fed to the answer generator. In our main experiments, the window size is 20, the redundancy and coarsening thresholds are $\lambda_{\text{red}}=0.6$ and $\lambda_{\text{coal}}=0.7$, and the recall depths are $k_{\text{sem}}=20$ and $k_{\text{lex}}=5$.

\subsection{Main Results and Analysis}
We evaluate \ours{} across multiple LLM backbones and compare against recent memory baselines, including LoCoMo~\cite{locomo}, ReadAgent~\cite{readagent}, MemoryBank~\cite{memorybank}, MemGPT~\cite{Packer2023MemGPTTL}, A-Mem~\cite{Xu2025AMEMAM}, LightMem~\cite{fang2025lightmem}, Mem0~\cite{mem0,Chhikara2025Mem0BP}, and SimpleMem~\cite{liu2026simplemem}. Table~\ref{tab:high_cap_results} reports the breakdown by question type (MultiHop, Temporal, OpenDomain, SingleHop) and the associated Token Cost.

\noindent \textbf{Overall accuracy.}
\ours{} attains the highest Average F1 for all evaluated backbones. On GPT-4o, \ours{} reaches 42.57 Average F1, compared to 39.06 for SimpleMem and 36.09 for Mem0. On GPT-4.1-mini, \ours{} reaches 46.30, compared to 43.24 for SimpleMem and 34.20 for Mem0. On Qwen3-Plus, \ours{} reaches 46.03, compared to 37.49 for SimpleMem and 35.85 for Mem0. These results indicate that the connected-evidence retrieval improves accuracy beyond flat retrieval under the same backbone.

\noindent \textbf{MultiHop and Temporal.}
MultiHop is where \ours{} shows its largest gains. On GPT-4o, \ours{} reaches 41.34 MultiHop F1, compared to 35.89 for SimpleMem and 35.13 for Mem0. On Qwen3-Plus, \ours{} reaches 42.17, compared to 33.74 for SimpleMem and 32.42 for Mem0. On GPT-4.1-mini, \ours{} reaches 44.24, compared to 43.46 for SimpleMem and 30.14 for Mem0. Temporal follows a similar trend. For example, on GPT-4o \ours{} reaches 57.94 Temporal F1, compared to 56.71 for SimpleMem and 52.38 for Mem0. These results support the design choice of connecting scattered evidence with bridge nodes and then conditioning generation on the resulting subgraph.
Temporal matters because it reflects timeline ordering and state updates across sessions. On Qwen3-Plus, \ours{} reaches 63.67 Temporal F1, compared to 50.87 for SimpleMem. On GPT-4.1-mini, \ours{} reaches 64.28, compared to 58.62 for SimpleMem. This aligns with conflict-aware coarsening for updates and time-aware graph completion, and it is reinforced by the generator's temporal-fidelity constraints.

\noindent \textbf{Token cost.}
On GPT-4o, \ours{} uses 497 tokens, compared to 550 for SimpleMem and 985 for Mem0. On Qwen3-Plus, \ours{} uses 460 tokens, compared to 583 for SimpleMem and 1,020 for Mem0. On GPT-4.1-mini, \ours{} uses 916 tokens, compared to 531 for SimpleMem. Overall, \ours{} keeps Token Cost below full-context baselines while improving accuracy, and the remaining variation reflects the node budget and how much intermediate evidence is needed for a query.

\begin{table*}[t]
  \centering
  \caption{Performance on the LoCoMo benchmark with High-Capability Models (GPT-4.1 series, GPT-4o, and Qwen3-Plus).}
  \label{tab:high_cap_results}
  \resizebox{0.95\textwidth}{!}{%
  \begin{tabular}{l|l|cc|cc|cc|cc|cc|r}
  \toprule
  \multirow{2}{*}{\textbf{Model}} & \multirow{2}{*}{\textbf{Method}} & \multicolumn{2}{c|}{\textbf{MultiHop}} & \multicolumn{2}{c|}{\textbf{Temporal}} & \multicolumn{2}{c|}{\textbf{OpenDomain}} & \multicolumn{2}{c|}{\textbf{SingleHop}} & \multicolumn{2}{c|}{\textbf{Average}} & \multicolumn{1}{c}{\textbf{Token}} \\
   & & \textbf{F1} & \textbf{BLEU} & \textbf{F1} & \textbf{BLEU} & \textbf{F1} & \textbf{BLEU} & \textbf{F1} & \textbf{BLEU} & \textbf{F1} & \textbf{BLEU} & \multicolumn{1}{c}{\textbf{Cost}} \\
  \midrule
  
  \multirow{8}{*}{\textbf{GPT-4.1-mini}} 
   & LoCoMo & 25.02 & 21.62 & 12.04 & 10.63 & 19.05 & 17.07 & 18.68 & 15.87 & 18.70  & 16.30 & 16,910 \\
   & ReadAgent & 6.48 & 5.6 & 5.31 & 4.23 & 7.66 & 6.62 & 9.18 & 7.91 & 7.16  & 6.09 & 643 \\
   & MemoryBank & 5.00 & 4.68 & 5.94 & 4.78 & 5.16 & 4.52 & 5.72 & 4.86 & 5.46  & 4.71 & \textbf{432} \\
   & MemGPT & 17.72 & 16.02 & 19.44 & 16.54 & 11.29 & 10.18 & 25.59 & 24.25 & 18.51  & 16.75 & 16,977 \\
   & A-Mem & 25.06 & 17.32 & 51.01 & 44.75 & 13.22 & 14.75 & 41.02 & 36.99 & 32.58  & 28.45 & 2,520 \\
   & LightMem & 24.96 & 21.66 & 20.55 & 18.39 & 19.21 & 17.68 & 33.79 & 29.66 & 24.63  & 21.85  & 612 \\
   & Mem0 & 30.14 & 27.62 & 48.91 & 44.82 & 16.43 & 14.94 & 41.3 & 36.17 & 34.20  & 30.89 & 973 \\
  & SimpleMem & 43.46 & \textbf{38.82} & 58.62 & 50.10 & 19.76 & 18.04 & 51.12 & 43.53 & 43.24  & 37.62 & 531 \\
  \rowcolor{gray!10} & \textbf{\ours{}} & \textbf{44.24} & 37.04 & \textbf{64.28} & \textbf{52.33} & \textbf{22.56} & \textbf{18.86} & \textbf{54.11} & \textbf{48.65} & \textbf{46.30}  & \textbf{39.22} & 916 \\
  \midrule
  
  \multirow{8}{*}{\textbf{GPT-4o}} 
   & LoCoMo & 28.00 & 18.47 & 9.09 & 5.78 & 16.47 & 14.80 & 61.56 & 54.19 & 28.78 & 23.31 & 16,910 \\
   & ReadAgent & 14.61 & 9.95 & 4.16 & 3.19 & 8.84 & 8.37 & 12.46 & 10.29 & 10.02 & 7.95 & 805 \\
   & MemoryBank & 6.49 & 4.69 & 2.47 & 2.43 & 6.43 & 5.30 & 8.28 & 7.10 & 5.92 & 4.88 & 569 \\
   & MemGPT & 30.36  & 22.83  & 17.29  & 13.18  & 12.24  & 11.87  & 40.16  & 36.35  & 25.01  & 21.06  & 16,987 \\
   & A-Mem & 32.86  & 23.76  & 39.41  & 31.23  & 17.10  & 15.84  & 44.43  & 38.97  & 33.45  & 27.45 & 1,216 \\
   & LightMem & 28.15 & 21.83 & 36.53 & 29.12 & 13.38 & 11.54 & 33.76  & 28.02  & 27.96  & 22.63 & 645 \\
   & Mem0 & 35.13 & 27.56 & 52.38 & 44.15 & 17.73 & 15.92 & 39.12 & 35.43 & 36.09  & 30.77 & 985 \\
  & SimpleMem & 35.89 & 32.83 & 56.71 & 20.57 & 18.23 & 16.34 & 45.41 & 39.25 & 39.06  & 27.25  & 550 \\
  \rowcolor{gray!10} & \textbf{\ours{}} & \textbf{41.34} & \textbf{34.49} & \textbf{57.94} & \textbf{46.34} & \textbf{25.02} & \textbf{22.93} & \textbf{45.97} & \textbf{40.11} & \textbf{42.57}  & \textbf{35.97} & \textbf{497} \\
  \midrule
  
  \multirow{8}{*}{\textbf{Qwen3-Plus}} 
   & LoCoMo & 24.15 & 18.94 & 16.57 & 13.28 & 11.81 & 10.58 & 38.58 & 28.16 & 22.78 & 17.74 & 16,910 \\
   & ReadAgent & 9.52 & 6.83 & 11.22 & 8.15 & 5.41 & 5.23 & 9.85 & 7.96 & 9.00 & 7.04 & 742 \\
   & MemoryBank & 5.25 & 4.94 & 1.77 & 6.26 & 5.88 & 6.00 & 6.90 & 5.57 & 4.95 & 5.69 & \textbf{302} \\
   & MemGPT & 25.80 & 17.50 & 24.10 & 18.50 & 9.50 & 7.80 & 40.20 & 42.10 & 24.90 & 21.48 & 16,958 \\
   & A-Mem & 26.50 & 19.80 & 46.10 & 35.10 & 11.90 & 11.50 & 43.80 & 36.50 & 32.08 & 25.73 & 1,427 \\
   & LightMem & 28.95 & 24.13 & 42.58 & 38.52 & 16.54 & 13.23 & 40.78 & 36.52 & 32.21 & 28.10 & 606 \\
   & Mem0 & 32.42 & 21.24 & 47.53 & 39.82 & 17.18 & 14.53 & 46.25 & 37.52 & 35.85  & 28.28 & 1,020 \\
  & SimpleMem & 33.74 & 29.04 & 50.87 & 43.31 & 18.41 & 16.24 & 46.94 & 38.16 & 37.49  & 31.69 & 583 \\
  \rowcolor{gray!10} & \textbf{\ours{}} & \textbf{42.17} & \textbf{36.06} & \textbf{63.67} & \textbf{49.84} & \textbf{24.46} & \textbf{22.92} & \textbf{53.80} & \textbf{48.71} & \textbf{46.03}  & \textbf{39.38} & 460 \\
  \bottomrule
  \end{tabular}
  }
  \end{table*}


\subsection{Efficiency Comparison}
\label{sec:efficiency}
We compare efficiency across memory systems along three axes: retrieval latency, retrieved context length (Token Cost), and end-to-end runtime.
We report \textbf{retrieval time} as the wall-clock time for the retriever to return evidence, excluding the final answer generation call.
\begin{table}[t]
\centering
\caption{Efficiency comparison on LoCoMo~\cite{locomo}. We report construction time, retrieval time, total time, and Average F1, here Baseline methods are based on GPT-4.1-mini.}
\label{tab:efficiency}
\resizebox{0.90\textwidth}{!}{
\begin{tabular}{l|c|c|c|c}
\toprule
\textbf{Method} & \textbf{Construction Time (s)} & \textbf{Retrieval Time (s)} & \textbf{Total Time(s)} & \textbf{Average F1} \\
\midrule
A-mem &5140.5s &796.7s &5937.2s& 32.58\\
Lightmem &97.8s &577.1s &675.9s &24.63\\
Mem0 &1350.9s &583.4s &1934.3s &34.20\\
SimpleMem &92.6s &388.3s &480.9s &43.24\\
\rowcolor{gray!10} \textbf{\ours{} w GPT-4o} & \textbf{38.0s} & {391.9s} & {429.9s} & {42.57}\\
\rowcolor{gray!10} \textbf{\ours{} w GPT-4.1-mini} & {113.5s} & \textbf{299.7s} & \textbf{413.2s} & \textbf{46.30}\\
\bottomrule
\end{tabular}
}
\end{table}
Table~\ref{tab:efficiency} reports runtime breakdown. Compared to Mem0, \ours{} reduces total time from 1934.3s to 429.9s (GPT-4o) and increases Average F1 from 34.20 to 42.57. Compared to SimpleMem, \ours{} reduces total time from 480.9s to 429.9s (GPT-4o), with Average F1 42.57 vs.\ 43.24. Under GPT-4.1-mini, \ours{} reaches 46.30 Average F1 at 413.2s total time.
Taken together, Table~\ref{tab:efficiency} shows that \ours{} reduces total runtime relative to iterative retrieval baselines while keeping accuracy competitive with strong structured-memory systems.

\subsection{Ablation Study}
\label{sec:ablation}
We ablate key components of \ours{} to quantify their contribution to accuracy and efficiency.
In the final version, we will report both (i) accuracy (Avg.\ F1; and optionally per-category F1) and (ii) efficiency (Token Cost / retrieval time) for each ablation.

\noindent \textbf{Impact of Entropy-Aware Gating.}
Removing gating reduces Average F1 from 42.57 to 41.68, a drop of 0.89. This indicates that gating helps efficiency and contributes modestly to accuracy under this setup.

\noindent \textbf{Impact of Conflict-Aware Coarsening.}
Removing coarsening reduces Average F1 from 42.57 to 41.42, a drop of 1.15. The per-category changes differ by subset. SingleHop increases from 45.97 to 47.11, while Temporal decreases from 57.94 to 53.93. This suggests coarsening trades redundancy reduction against preserving fine-grained updates.

\noindent \textbf{Impact of Bridge Discovery.}
Removing bridge discovery reduces Average F1 from 42.57 to 35.57, a drop of 7.00. The largest drops are on MultiHop, which goes from 41.34 to 35.95, and Temporal, which goes from 57.94 to 44.64. This supports the use of bridge nodes to connect evidence that is missed by initial hybrid recall.

\noindent \textbf{Impact of Topology-Aware Reasoning.}
Removing topology-aware reasoning reduces Average F1 from 42.57 to 36.27, a drop of 6.30. MultiHop decreases from 41.34 to 33.67 and Temporal decreases from 57.94 to 45.66. This indicates that passing the same facts without structural guidance is not sufficient for multi-evidence questions.
Overall, Table~\ref{tab:ablation_by_type} shows that bridge discovery and topology-aware reasoning are the two dominant components for MultiHop and Temporal accuracy under GPT-4o.

\begin{table*}[t]
\centering
\caption{Ablation study by question type on LoCoMo~\cite{locomo} with GPT-4o as the backbone.}
\label{tab:ablation_by_type}
\resizebox{0.95\textwidth}{!}{%
\begin{tabular}{l|cc|cc|cc|cc|c}
\toprule
\multirow{2}{*}{\textbf{Configuration}} & \multicolumn{2}{c|}{\textbf{MultiHop}} & \multicolumn{2}{c|}{\textbf{Temporal}} & \multicolumn{2}{c|}{\textbf{OpenDomain}} & \multicolumn{2}{c|}{\textbf{SingleHop}} & \textbf{Avg.\ F1}  \\
 & \textbf{F1} & \textbf{BLEU} & \textbf{F1} & \textbf{BLEU} & \textbf{F1} & \textbf{BLEU} & \textbf{F1} & \textbf{BLEU} &   \\
\midrule
\rowcolor{gray!10} \textbf{Full \ours{}} & \textbf{41.34} & \textbf{34.49} & \textbf{57.94} & \textbf{46.34} & \textbf{25.02} & \textbf{22.93} & \textbf{45.97} & \textbf{40.11} & \textbf{42.57} \\
\midrule
\quad w/o Entropy-aware gating  & 41.12 & 33.03 & 55.72 & 44.28 & 24.32 & 21.64 & 45.58 & 39.81 & 41.68 \\
\quad w/o Coarsening & 40.48 & 33.06 & 53.93 & 42.95 & 24.15 & 22.13 & 47.11 & 41.66 & 41.42 \\
\quad w/o bridge discovery & 35.95 & 25.55 & 44.64 & 32.25 & 20.51 & 15.17 & 41.19 & 35.27 & 35.57 \\
\quad w/o topology-aware reasoning & 33.67 & 24.79 & 45.66 & 33.56 & 23.97 & 18.66 & 41.79 & 35.77 & 36.27 \\
\bottomrule
\end{tabular}
}
\end{table*}

\begin{table}[t]
\centering
\caption{Comparison of retrieval depth sensitivity with GPT-4o as the backbone.}
\label{tab:combined_k_sensitivity}

\begin{subtable}[b]{0.495\textwidth}
\centering
\caption{$k_{\text{sem}}$}
\begin{tabular}{c|c|c}
\toprule
$k$ & Avg. F1 & Retrieval Time (s)  \\
\midrule
1  & 27.51 & 265.3s \\
10  & 39.59 & 392.3s \\
\rowcolor{gray!10} \textbf{20}  &  \textbf{42.57} &  \textbf{391.9s} \\
30 & 40.27 & 276.8s \\
\bottomrule
\end{tabular}
\end{subtable}
\hfill
\begin{subtable}[b]{0.495\textwidth}
\centering
\caption{$k_{\text{lex}}$}
\begin{tabular}{c|c|c}
\toprule
$k$ & Avg. F1 & Retrieval Time (s)  \\
\midrule
1  & 42.25 & 385.9s \\
3  & 42.40 & 392.7s \\
\rowcolor{gray!10} \textbf{5}  &  \textbf{42.57} &  \textbf{391.9s} \\
10 & 41.16 & 288.7s \\
\bottomrule
\end{tabular}
\end{subtable}

\end{table}

Table~\ref{tab:combined_k_sensitivity} varies retrieval depth. Increasing $k_{\text{sem}}$ improves Average F1 from 27.51 at $k{=}1$ to 42.57 at $k{=}20$. Varying $k_{\text{lex}}$ yields smaller changes in Average F1. Average F1 is 42.25 at $k{=}1$ and 42.57 at $k{=}5$. Retrieval time is not monotonic in $k$ because node limiting and graph construction can dominate runtime.

\section{Conclusion}
In this paper, we presented AriadneMem, a structured memory system designed to thread the complex maze of lifelong memory for LLM agents. By decoupling the memory pipeline into an offline construction phase and an online reasoning phase, we effectively address the persistent challenges of disconnected evidence and state updates in long-term dialogues.
AriadneMem leverages a conflict-aware coarsening mechanism to maintain a dynamic, evolving world model through structured temporal edges. By introducing bridge discovery and DFS-based path mining, we transform the disjointed fragments of long-horizon contexts into coherent reasoning chains, effectively resolving the challenges of multi-hop information retrieval in disconnected evidence scenarios.

In summary, AriadneMem achieves a superior balance of efficiency and accuracy. By replacing iterative planning with structural graph traversal, it reduces runtime by 77.8\% while boosting Multi-Hop F1 by 15.2\%. With its compact token footprint, AriadneMem provides a fast, accurate, and scalable memory foundation for long-horizon LLM agents.

\bibliography{colm2026_conference}
\bibliographystyle{colm2026_conference}



\end{document}